\documentclass[12pt]{article}  
\usepackage{color,amssymb,amsthm,amsmath,amsxtra,amsfonts,breqn,
fullpage,graphicx,hyperref,float,appendix,mathtools,eqparbox, 
lscape,pdfpages,comment,framed,chngpage}

\usepackage{multirow}
\usepackage{booktabs}

\usepackage[most]{tcolorbox}

\usepackage[latin1]{inputenc}
\usepackage{tikz}
\usetikzlibrary{shapes,arrows}

\newcommand{\overbar}[1]{\mkern 1.5mu\overline{\mkern-1.5mu#1\mkern-1.5mu}\mkern 1.5mu}

\usepackage{fancyhdr}
\usepackage{xcolor}
\definecolor{dark-red}{rgb}{0.4,0.15,0.15}
\definecolor{dark-blue}{rgb}{0,0,0.45}
\definecolor{dark-olive-green}{rgb}{0.333,0.42,0.184}
\definecolor{magenta}{rgb}{1.,0.,1.}
\hypersetup{colorlinks, linkcolor={dark-blue},citecolor={blue}, urlcolor={blue}}

\numberwithin{equation}{section}
\usepackage{pdflscape}

\definecolor{gold}{rgb}{0.85,0.66,0.0}
\definecolor{lightergray}{rgb}{0.954543,0.961501,0.949737}

\theoremstyle{definition}

\usepackage{etoolbox}
\makeatletter
\providecommand{\institute}[1]{
  \apptocmd{\@author}{\end{tabular}
    \par
    \begin{tabular}[t]{c}
    #1}{}{}
}
\makeatother

\usepackage[ddmmyyyy,hhmmss]{datetime}

\font\eightss=cmssq8
\font\eightssi=cmssqi8
\newcommand\quoteAuthorDate[3]{\begingroup
  \baselineskip 10pt
  \parfillskip 0pt
  \interlinepenalty 10000 
  \leftskip 0pt plus 40pc minus \parindent
  \let\rm=\eightss
  \let\sl=\eightssi
  \everypar{\sl}#1\par
  \nobreak\smallskip
  \noindent\rm--- #2\unskip\enspace(#3)\par
  \endgroup}

\begin{document}

\title{Integer Factorisation, Fermat \& Machine Learning on a Classical Computer}
\author{Sam Blake\\ {\small\texttt{samuel [dot] thomas [dot] blake [at] gmail [dot] com} }}
\date{\today}
\maketitle

\bigskip

\begin{abstract}
In this paper we describe a deep learning--based probabilistic algorithm for integer factorisation. We use Lawrence's extension of Fermat's factorisation algorithm to reduce the integer factorisation problem to a binary classification problem. To address the classification problem, based on the ease of generating large pseudo--random primes, a corpus of training data, as large as needed, is synthetically generated. We will introduce the algorithm, summarise some experiments, analyse where these experiments fall short, and finally put out a call to others to reproduce, verify and see if this approach can be improved to a point where it becomes a practical, scalable factorisation algorithm. 
\end{abstract}

\bigskip



\section{Introduction}
Factoring integers into primes is notoriously difficult. If you're reading this paper, chances are you already know that, but it's worth stating. The difficulty of integer factorisation is exploited to form the backbone of the ubiquitous RSA public--key cryptosystem\cite{rsa}. \\

At present, no practical, polynomial time algorithm for integer factorisation exists which runs on a classical computer. However, there is no known theoretical reason why integer factorisation should be difficult, even for an algorithm which runs on a classical computer\cite{cohen}. \\

It is conjectured that in the future quantum computers will be able to factor large, RSA--type semiprimes using Shor's algorithm and consequently break RSA encryption\cite{shor1994}. Of course, we do not know exactly when this will happen, if ever. Nation states are betting on such a breakthrough and storing vast amounts of encrypted data in the hope that one day they will be able to decrypt it using such a breakthrough. This has been referred to as a Harvest Now, Decrypt Later (HNDL) attack\cite{hndl}. \\

We can track the (known) state of the art in integer factorisation using the RSA Challenge Numbers\cite{rsa_challenge_numbers}. These semiprimes were constructed in 1991 to encourage research into integer factorisation and track the practical difficulty of factoring large semiprimes and consequently cracking RSA keys used in contemporary cryptography\cite{kaliski1991}. RSA--250, a 250 decimal digit semiprime with no known weaknesses, was factored in 2020 by a team lead by Paul Zimmermann\cite{rsa_250_factored}. At present, the factorisation of RSA--260 has not been announced, nor has the factorisations of any larger RSA Challenge Numbers. \\

The oldest and best known algorithm for integer factorisation is trial division, which begins by dividing--out all powers of 2, powers of 3, then divides--out odd numbers of the form $6k\pm1$ for $k$ from 1 to $\lfloor\sqrt{\mathcal{N}}\rfloor$. Trial division was first described by Fibonacci in his book \textit{Liber Abaci} in 1202. There are many modern factorisation algorithms including the continued fraction factorisation algorithm by Lehmer and Powers\cite{lehman1974}, Pollard's \textit{rho} and $p-1$ algorithms\cite{pollard1974}\cite{pollard1975}, Shanks SQUFOF (square forms of factoring) algorithm\cite{shanks}, Dixon's \textit{random squares method}\cite{dixon1981}, Pomerance's quadratic sieve factoring algorithm\cite{pomerance1984}, Lenstra's elliptic curve factorisation algorithm\cite{lenstra1987}, Silverman's multiple polynomial quadratic sieve\cite{silverman1987}, and most recently Hart's \textit{one line factoring algorithm}\cite{hart2012}. An excellent history of integer factorisation is given by Wagstaff\cite{wagstaff2013}\cite{wagstaff2021}.\\

While the integer factorisation problem is not known to be NP--hard, there is a relatively recent precedent for using deep learning to find approximate solutions to NP--hard problems. The AlphaFold 2 algorithm, developed by the Google DeepMind team in 2021, is a deep learning-based method primarily designed for predicting protein structures\cite{alphafold2021}. The success of AlphaFold in predicting protein structures stems from its ability to encode complex spatial relationships between amino acids and constructing complex deep learning models to capture intricate patterns and representations from training data.

\section{Fermat's Factorisation Algorithm}

Devised by the ingenious Pierre de Fermat in 1643 is a factorisation algorithm which is based on representing an odd integer, $\mathcal{N}$, as the difference of two squares
$$\mathcal{N} = a^2 - b^2.$$
If such a representation is found, then we have the (algebraic) factorisation 
$$\mathcal{N} = (a-b)(a+b).$$
If neither factor is 1, then we have a non--trivial factorisation of $\mathcal{N}$. Every odd number, $\mathcal{N} = p\,q$, possesses such a representation, as 
$$\mathcal{N} = \left(\frac{p+q}{2}\right)^2 - \left(\frac{p-q}{2}\right)^2.$$

In its simplest form, Fermat's algorithm starts with $a = \left\lfloor\sqrt{\mathcal{N}}\right\rfloor$ and checks if $a^2 - \mathcal{N}$ is a perfect square, $b^2$, if not $a$ is incremented until a perfect square is found. In Python 3, we have 

\scriptsize
\begin{tcolorbox}[colframe=white,breakable]
\begin{verbatim}
%pip install gmpy2 
import gmpy2
from gmpy2 import mpz, mpq, mpfr

def factor_fermat(n, max_iter = 65536):
  """Fermat's factorisation method (using GMP for fast bignum arithmetic.)"""

  a = gmpy2.isqrt(n)
  b = a**2 - n
    
  n_iter = 0
  while not gmpy2.is_square(b):
    a += 1
    b = a**2 - n
    n_iter += 1
    if n_iter > max_iter:
      print(f'max_iter of {max_iter} exceeded.')
      return mpz(1)

  print(f'n_iterations = {n_iter}')
  return a - gmpy2.isqrt(b)
\end{verbatim}
\end{tcolorbox}
\normalsize

Various improvements to Fermat's algorithm have been proposed, a nice summary is found in Bahig\cite{bahig2020}. However, the version of Fermat's algorithm given above is sufficient for our (purely illustrative) purposes.\\

If $p$ and $q$ are primes, and $\mathcal{N} = p\,q$, then Fermat's algorithm is \textit{quite efficient} if $p/q$ is near $1$ (hence $p,q$ are close to $\sqrt{\mathcal{N}}$), but the number of trials required quickly grows if $p/q$ is not close to $1$\cite{lehman1974}. \\

For example, consider primes $p,q$ and semiprime, $\mathcal{N} = p\,q$, generated in Python 3 by computing
\scriptsize
\begin{tcolorbox}[colframe=white,breakable]
\begin{verbatim}
p = gmpy2.next_prime(2**n_bits + 2**n_lsb_bits)
q = gmpy2.next_prime(2**n_bits)
N = p*q
\end{verbatim}
\end{tcolorbox}
\normalsize

where $p,q$ are \verb~n_bits~--bit primes and \verb~n_lsb_bits~ is used to modify the difference between $p$ and $q$. As a way to empirically estimate how \textit{close} $p/q$ must be to $1$ to practically factor $\mathcal{N}$, we can compute the number of iterations Fermat's algorithm takes to compute a factor of $\mathcal{N}$ as a function of \verb~n_lsb_bits/n_bits~. This is summarised in the following plot for \verb~n_bits~ $\in [100, 300, 500, 700, 900]$:

\begin{figure}[H]
\centering
\includegraphics[width=0.5\textwidth]{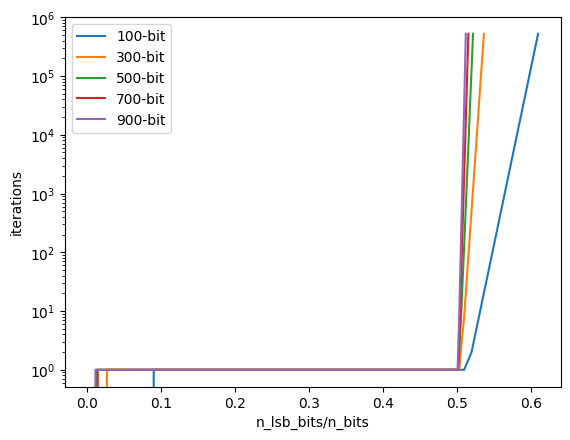}
\caption{\small A plot of the number of iterations for Fermat's factorisation algorithm.}
\end{figure}

Thus, once \verb~n_lsb_bits/n_bits~ $> 0.5$ the number of iterations of Fermat's algorithm increases exponentially. As a concrete example, when \verb~n_lsb_bits/n_bits~ $=0.4$, Fermat's algorithm requires only 1 iteration to compute the following prime factorisation
{\footnotesize
\begin{multline*}
1606938044258990276935758667842587029774996746028337250487979 = \\
126765060022822940\textcolor{gray}{2596214833343} \times 126765060022822940\textcolor{gray}{1496703205653},
\end{multline*}
}
and when \verb~n_lsb_bits/n_bits~ $=0.6$, it requires 131\,072 iterations to compute the following prime factorisation 
{\footnotesize
\begin{multline*}
1606938044260451777179292995662479178227815585735176483777629 = \\
126765060022\textcolor{gray}{9382323001310052393} \times 126765060022\textcolor{gray}{8229401496703205653}.
\end{multline*}
}

In general, Fermat's factorisation algorithm will factor $\mathcal{N}$ quickly when $p$ is within $\mathcal{O}\left(\sqrt[4]{\mathcal{N}}\right)$ of $\sqrt{\mathcal{N}}$. \\

In 1895, Lawrence\cite{lawrence1895} extended Fermat's algorithm, when $p/q$ is not close to $1$, but we have $u,v$ such that $u/v$ is \textit{sufficiently close} to $p/q$, then we can use Fermat's algorithm to factor $u\,v\,\mathcal{N}$. If one factor of $u\,v\,\mathcal{N}$ is $m$, then $\gcd(m,\mathcal{N})$ is a factor of $\mathcal{N}$. \\

Lawrence's extension of Fermat's algorithm is trivial to implement. In Python 3, we have 
\scriptsize
\begin{tcolorbox}[colframe=white,breakable]
\begin{verbatim}
def factor_lawrence(n, u_v_ratio, max_iter = 100000):
  """Lawrence's extension to Fermat's factoring algorithm."""
  u,v = u_v_ratio.numerator, u_v_ratio.denominator
  return gmpy2.gcd(factor_fermat(u*v*n, max_iter),n)
\end{verbatim}
\end{tcolorbox}
\normalsize

As an example, given primes $$p=33059500175075655435169$$ and $$q = 22642302873041910393781,$$
with $$\mathcal{N} = p\,q = 748543215795445052722625573101291605706283989.$$ We cannot immediately factor $\mathcal{N}$ using Fermat's algorithm, however if by some means we have the approximation 
$$p/q \approx 210381/144089,$$ 
then we can quickly factor 
$$\mathcal{M} = 210381\times144089\times \mathcal{N}$$ 
using Lawrence's extension of Fermat's factorisation algorithm, as 
$$\mathcal{M} = 4763510320729953133276053301^2-3477017277987260^2.$$
This only took 1268 iterations, as $4763510320729953133276053301$ is close to $\left\lfloor \sqrt{\mathcal{M}} \right\rfloor$. Then 
\begin{multline*}
\mathcal{M} = (4763510320729953133276053301-3477017277987260) \times \\
(4763510320729953133276053301+3477017277987260)\\
= 4763510320726476115998066041 \times 4763510320733430150554040561,
\end{multline*}
and we can now obtain the prime factorisation of $\mathcal{N}$ as follows $$\gcd(4763510320726476115998066041,\mathcal{N}) = 33059500175075655435169$$ and $$\gcd(4763510320733430150554040561,\mathcal{N}) = 22642302873041910393781.$$

\section{Binary Classification with Deep Learning}\label{binary_classification_section}

Binary classification is a type of supervised machine learning task in which an algorithm is trained to predict one of two possible outcomes or classes for a given input. \\

In binary classification, the algorithm is trained using a labeled dataset that contains examples of inputs and their corresponding class labels. The goal is to learn a model that can accurately predict the class label of new, unseen inputs. \\

The algorithm we present in this paper was designed around the performance of the following binary classification experiment. \\

Our goal is to create a binary classifier which given an $n$--bit semiprime, $\mathcal{N} = p\,q$, can decide if $R_{\text{min}} < p/q < R_{\text{max}}$ for some user--defined interval $\left(R_{\text{min}},R_{\text{max}}\right)$. This classification is blind, that is, the classifier is not given access to the prime factors $p$ and $q$. The binary classifier we chose was a 3 layer densely--connected neural network, which was constructed using the Keras deep learning library\cite{keras}:

\scriptsize
\begin{tcolorbox}[colframe=white,breakable]
\begin{verbatim}
from tensorflow.keras.models import Sequential
from tensorflow.keras.layers import Dense, Dropout
from tensorflow.keras import regularizers

def create_baby_model():
    # Create model
    model = Sequential()

    model.add(Dense(n_semiprime_bits//4, \
      kernel_regularizer=regularizers.l2(0.0005), \
      input_shape=(n_semiprime_bits,), \
      activation='relu'))

    model.add(Dropout(0.2))
    model.add(Dense(n_semiprime_bits//4, \
      kernel_regularizer=regularizers.l2(0.0005), \
      activation='relu'))

    model.add(Dropout(0.2))
    model.add(Dense(1, activation='sigmoid'))

    # Compile model
    model.compile(loss='binary_crossentropy', optimizer='adam', metrics=['accuracy'])
    return model
\end{verbatim}
\end{tcolorbox}
\normalsize

We consider this model simple, relative to more modern machine learning techniques, including LSTM networks\cite{lstm1997} and transformers\cite{vaswani2017}. \\

Generating a large training dataset of $[\mathcal{N}, R_{\text{min}} < p/q < R_{\text{max}}]$--pairs is fast, as large pseudo--random primes are easily generated using the \texttt{random} module from the Python standard library and the GNU MP bignum library, \texttt{gmpy2}\cite{gmp}. The Python 3 code for generating a single random $n$--bit semiprime where $p/q$ is within a user--specified interval, $R_{\text{min}} < p/q < R_{\text{max}}$ is given by 
\scriptsize
\begin{tcolorbox}[colframe=white,breakable]
\begin{verbatim}
import random

def random_prime_pair(n_bits, ratio_interval):
  """random_prime_pair returns a semiprime and a pair of random primes with ratio \
  randomly within the ratio_interval, which is given by [mpq(n1,d1), mpq(n2,d2)]. \
  The product of the two semiprimes will be n_bits bits."""
    
  interval_lower, interval_upper = ratio_interval
  assert interval_lower < interval_upper
    
  while True:
    rn = random.randint(0,2**64 - 1)
    rd = random.randint(1,2**64 - 1)
    
    if rn > rd:
      rn,rd = rd,rn

    rq = mpq(rn, rd) # 0 <= rq <= 1
    r = interval_lower + (interval_upper - interval_lower)*rq
    r_nbits = int(gmpy2.floor(gmpy2.log2(r)) + 1)
    rmin, rmax = 2**(n_bits//2 - r_nbits//2 - 1), 2**(n_bits//2 - r_nbits//2)
    rint_q = mpz(random.randint(rmin,rmax))
    rint_p, _ = gmpy2.t_divmod(r.numerator*rint_q, r.denominator)
    
    p = gmpy2.next_prime(rint_p)
    q = gmpy2.next_prime(rint_q)
    N = gmpy2.mul(p,q)
    if len(N.digits(2)) == n_bits:
      break
    
  if p < q:
    p,q = q,p

  return N,p,q
\end{verbatim}
\end{tcolorbox}
\normalsize

For example, below we use \texttt{random_prime_pair} to generate a pseudo--random 426--bit semiprime, $\mathcal{N} = p \, q$, with $1 < p/q < 2$:
\scriptsize
\begin{tcolorbox}[colframe=white,breakable]
\begin{verbatim}
>>> N,p,q = random_prime_pair(426, [1, 2])
N, p, q, p/q, len(p.digits(2)), len(q.digits(2)), len(N.digits(2))
\end{verbatim}
\end{tcolorbox}
\normalsize

\scriptsize
\begin{tcolorbox}[colframe=white,breakable]
\begin{verbatim}
(mpz(15663643342929165851763765513691864638402240195390267858012917780570810151199\
7625415213950022665183877053763783758440589187403561),
 mpz(16748329550089039633739413650829723130328207468807461595876202133),
 mpz(9352361557063995535602420995718150617295463074650159574587938117),
 mpfr('1.7908128816340239'),
 214,
 213,
 426)
\end{verbatim}
\end{tcolorbox}
\normalsize

We require our training dataset to be balanced with respect to our classification $R_{\text{min}} < p/q < R_{\text{max}}$, and we also require the example semiprimes not within our interval $(R_{\text{min}}, R_{\text{max}})$, to be within $R_{\text{min}} - \Delta < p/q < R_{\text{max}} + \Delta$, where $\Delta = (R_{\text{max}} - R_{\text{min}})/2$. The Python 3 code for generating the training data is given below.
\scriptsize
\begin{tcolorbox}[colframe=white,breakable]
\begin{verbatim}
import tqdm.notebook as tq

def generate_training_semiprimes(n_semiprime_bits, \
                                 min_ratio, max_ratio, ratio_diff_scale, \
                                 n_training_samples):
  """generate_training_semiprimes returns a list of /n_training_samples/ pairs 
  of the form [/semiprime/, p/q-decision], used for subsequent training for 
  the binary classification problem."""

  diff = max_ratio - min_ratio 
  diff *= ratio_diff_scale
  min_extended_ratio_mpq = max(1, gmpy2.f2q(min_ratio - diff))
  max_extended_ratio_mpq = max(1, gmpy2.f2q(max_ratio + diff))
  pq_ratio_interval = [min_extended_ratio_mpq, max_extended_ratio_mpq]

  assert min_extended_ratio_mpq < max_extended_ratio_mpq

  n_inside = 0
  n_outside = 0

  semiprimes_inside = dict()
  semiprimes_outside = dict()
    
  pbar = tq.tqdm(total = n_training_samples)

  while n_inside + n_outside < n_training_samples:

    # Generate random prime pair within a specified p/q-ratio interval. 
    N,p,q = random_prime_pair(n_semiprime_bits, pq_ratio_interval)            
    ratio = gmpy2.div(p,q)
    N = str(N)
     
    if n_inside <= n_training_samples//2 and min_ratio < ratio < max_ratio:
      pbar.update(1)
      n_inside += 1
      semiprimes_inside[N] = True

    if n_outside <= n_training_samples//2 and not min_ratio < ratio < max_ratio \
    and min_ratio - diff < ratio < max_ratio + diff:
      pbar.update(1)
      n_outside += 1
      semiprimes_outside[N] = True
  pbar.close()

  training_data = []
  for semiprime in semiprimes_inside.keys():
    training_data.append([semiprime, 1])

  for semiprime in semiprimes_outside.keys():
    training_data.append([semiprime, 0])
    
  random.shuffle(training_data)
  return training_data
\end{verbatim}
\end{tcolorbox}
\normalsize

Now we have the necessary code to generate our training data. As an example we generate 1 million, 426--bit semiprimes and test if $2 < p/q < 3$.

\scriptsize
\begin{tcolorbox}[colframe=white,breakable]
\begin{verbatim}
# 1 million 426-bit semiprimes, with p/q ratio in [2,3]

import pandas as pd

min_ratio = 2
max_ratio = 3
ratio_diff_scale = mpq(1,2)
n_semiprime_bits = 426
n_train = 10**6

training_data = generate_training_semiprimes(\
  n_semiprime_bits = n_semiprime_bits, \
  min_ratio = min_ratio, \
  max_ratio = max_ratio, \
  ratio_diff_scale = ratio_diff_scale, \
  n_training_samples = n_train)

# Write training data to file. 

file_name = f'training_data_{n_semiprime_bits}_ratio_{min_ratio}_{max_ratio}.h5'
df_training_data = pd.DataFrame(training_data)

if exists(file_name):
  df_training_data.to_hdf(
    file_name, 
    key='semiprimes', 
    append = True, 
    mode = 'r+', 
    format = 'table')
else:
  df_training_data.to_hdf(
    file_name, 
    min_itemsize = n_semiprime_bits, 
    key = 'semiprimes', 
    format = 'table')

training_data = pd.read_hdf(file_name)
training_data[0] = training_data[0].apply(mpz)
training_data = training_data.values
n_train = training_data.shape[0]
n_train
\end{verbatim}
\end{tcolorbox}
\normalsize

What features should we construct from our training data? At present, we use a simple single feature model, which is the binary representation of the semiprimes. Firstly, we need some code to do the base conversion and reshape our training data into a format usable by Keras:

\scriptsize
\begin{tcolorbox}[colframe=white,breakable]
\begin{verbatim}
def rat_base(n, b):
  """rat_base computes the (rational) base b representation of the positive 
  integer n, where b is a gmpy2 mpq object (a rational number)."""
    
  if type(b) is int or type(b) is mpz:
    return [int(k,b) for k in mpz(n).digits(b)]
  elif type(b) is float:
    b = gmpy2.f2q(b)

  if type(b) is not mpq:
    print('ERROR: type(b) == mpq.')

  if b < 1:
    print('ERROR: b > 1.')
    return n
    
  if n < 0:
    print('ERROR: n > 0')
    return n
    
  if n == 0:
    return [0]
    
  m = n
  base_rep = []
  while m > 0:
    d = gmpy2.f_mod(m, b.numerator)
    m = gmpy2.f_div(m, b.numerator)*b.denominator
    base_rep.append(int(d))
    
  return base_rep[::-1]
\end{verbatim}
\end{tcolorbox}
\normalsize

\scriptsize
\begin{tcolorbox}[colframe=white,breakable]
\begin{verbatim}
def pad_left(lst, n):
    """Makes a list of length n by padding with zeros on the left."""
    return [0]*(n - len(lst)) + lst
\end{verbatim}
\end{tcolorbox}
\normalsize

\scriptsize
\begin{tcolorbox}[colframe=white,breakable]
\begin{verbatim}
def reshape_training_data(training_data, base):
    """Reshapes the training data into a form suitable for use in Keras."""

    n_samples = len(training_data)
    max_len = len(rat_base(max(training_data[:,0]), base))

    X = np.zeros((n_samples, max_len), dtype = np.float32)
    for k, (train_d,_) in tq.tqdm(enumerate(training_data), total = n_samples):
        X[k,:] = pad_left(rat_base(train_d, base), max_len)

    Y = np.array([classification_d for _,classification_d in training_data])

    return X, Y
\end{verbatim}
\end{tcolorbox}
\normalsize

Now we have the code required to preprocess our training data. We construct the training data, using a $2/3$ train, $1/3$ test split of the training data: 

\scriptsize
\begin{tcolorbox}[colframe=white,breakable]
\begin{verbatim}
base = 2
n_semiprime_bits = 426
n_semiprime_bits_base_N = len(rat_base(max(training_data[:,0]), base))

# Convert training data to base. 
X, Y = reshape_training_data(training_data, base)
X_train, X_test, y_train, y_test = train_test_split(\
  X, Y, test_size = 0.33, shuffle = False)
\end{verbatim}
\end{tcolorbox}
\normalsize

And finally we can now train our model: 

\scriptsize
\begin{tcolorbox}[colframe=white,breakable]
\begin{verbatim}
# Fit model to training data.
es = EarlyStopping(monitor='accuracy', mode = 'max', restore_best_weights = True, \
  min_delta = 0.001, patience = 25, verbose = 1)

mc = ModelCheckpoint(\
  f'baby_model_n_semiprime_bits{n_semiprime_bits}_ratio_{min_ratio}_{max_ratio}_base_{base}.h5', \
  monitor = 'accuracy', mode = 'min', save_best_only = True)

estimator = KerasClassifier(model=create_baby_model, epochs=1000, \
  batch_size=100_000, verbose=1, callbacks = [es, mc])

fitted_model = estimator.fit(X_train, y_train)
\end{verbatim}
\end{tcolorbox}
\normalsize

The final few lines from the training process were:

\scriptsize
\begin{tcolorbox}[colframe=white,breakable]
\begin{verbatim}
Epoch 130/1000
7/7 [==============================] - 0s 43ms/step - loss: 0.5319 - accuracy: 0.7196
Epoch 131/1000
7/7 [==============================] - ETA: 0s - loss: 0.5314 - accuracy: 0.7200
Restoring model weights from the end of the best epoch: 106.
7/7 [==============================] - 0s 43ms/step - loss: 0.5314 - accuracy: 0.7200
Epoch 131: early stopping
\end{verbatim}
\end{tcolorbox}
\normalsize

So the best in--sample accuracy we achieved was 0.72. Let's check the out--of--sample performance: 

\scriptsize
\begin{tcolorbox}[colframe=white,breakable]
\begin{verbatim}
# Predict on test data. 
yhat = fitted_model.predict(X_test)
acc = accuracy_score(y_test, yhat)
print(f'accuracy = {acc:.4f}')
\end{verbatim}
\end{tcolorbox}
\normalsize

\scriptsize
\begin{tcolorbox}[colframe=white,breakable]
\begin{verbatim}
4/4 [==============================] - 0s 31ms/step
accuracy = 0.7164
\end{verbatim}
\end{tcolorbox}
\normalsize

An out--of--sample accuracy of 0.72 from such a simple model was unexpected. The corresponding out--of--sample confusion matrix is given below:

\scriptsize
\begin{tcolorbox}[colframe=white,breakable]
\begin{verbatim}
from sklearn.metrics import confusion_matrix
cm = confusion_matrix(y_test,yhat)
cm/np.sum(cm)
\end{verbatim}
\end{tcolorbox}
\normalsize
\medskip

\begin{center}
\begin{tabular}{@{}cc|cc@{}}
\multicolumn{1}{c}{} &\multicolumn{1}{c}{} &\multicolumn{2}{c}{\footnotesize\textsf{Predicted}} \\ 
\multicolumn{1}{c}{} & 
\multicolumn{1}{c|}{} & 
\multicolumn{1}{c}{T} & 
\multicolumn{1}{c}{F} \\ 
\cline{2-4}
\multirow[c]{2}{*}{\rotatebox[origin=tr]{90}{\footnotesize\textsf{Actual}}}
& T  & 0.273 & 0.227   \\[1.5ex]
& F  & 0.056 & 0.444 \\ 
\cline{2-4}
\end{tabular}
\end{center}

\medskip

From the confusion matrix we see the model rarely reports a false negative. \\

In this experiment we are assuming there isn't one or more (embarrassing) coding bugs which have skewed the model results, nor a bias accidentally introduced in the construction of the training data, nor some feature of semiprimes where $p/q$ is easily estimated from $\mathcal{N}$ which is unknown to the author. \\

\section{The Algorithm}

We previously showed that factoring $\mathcal{N} = p\,q$ can be achieved providing we have an approximation, $u/v$ which is sufficiently close to $p/q$. For practical purposes, we will define \textit{sufficiently close} as the Lawrence extension of Fermat's algorithm factoring $u\,v\,\mathcal{N}$ within \verb~max_iter~ iterations. We set \verb~max_iter~ to 100\,000, so attempting to factor $u\,v\,\mathcal{N}$ using \texttt{factor_lawrence} takes around a second for contemporary RSA--sized semiprimes. \\

We also previously showed the results of a simple experiment where a machine learning--based model estimated with reasonable accuracy if $R_{\text{min}} < p/q < R_{\text{max}}$, for some large semiprime, $\mathcal{N} = p\,q$. \\

We use these two results to construct the following stochastic binary search--based algorithm for integer factorisation. Given an $n$--bit semiprime, $\mathcal{N}$, a user--defined initial interval $[R_{\text{min}}, R_{\text{max}}]$, the number of training examples, $N_{\text{train}}$, and the minimum classification probability, $p_{\text{min}}$, we summarise the algorithm as follows.

\bigskip

\begin{tcolorbox}[colframe=white,breakable]
\noindent\texttt{factor_ml_based_binary_search}$\left(\mathcal{N}, [R_{\text{min}}, R_{\text{max}}], N_{\text{train}}, p_{\text{min}}\right)$:
\begin{enumerate}
\item Let $c = \left(R_{\text{max}} - R_{\text{min}}\right)/2$ be the bisection of the interval $[R_{\text{min}}, R_{\text{max}}]$. 
\item If $\texttt{factor_lawrence}(\mathcal{N}, c)$ returns a non--trivial factor, $p$ of $\mathcal{N}$, then \textbf{return} $p$. 
\item Denote $[R_{\text{min}}, c]$ and $[c, R_{\text{max}}]$ as the \textit{lower} and \textit{upper} intervals respectively. 
\item Construct two training datasets, each of size $N_{\text{train}}$, $n$--bit pseudo--random semiprimes for the lower and upper intervals. 
\item Train binary classification models for the upper and lower interval training datasets. Let $p_{\text{lower}}$ and $p_{\text{upper}}$ be the out--of--sample classification probability for the lower and upper training datasets. 
\item If $p_{\text{lower}}$ or $p_{\text{upper}}$ is less than $p_{\text{min}}$, then \textbf{stop} as the model has lost predictive capability.
\item Let $\mathcal{C}_{\text{lower}}$ and $\mathcal{C}_{\text{upper}}$ be the classifications of $\mathcal{N}$ for the models trained on the lower and upper intervals.
\item If both $\mathcal{C}_{\text{lower}}$ and $\mathcal{C}_{\text{upper}}$ are \texttt{False}, then \textbf{stop}, as the models have reached a conflict. 
\item If $\mathcal{C}_{\text{lower}}$ and $\mathcal{C}_{\text{upper}}$ do not conflict, then use this classification to update the interval $[R_{\text{min}}, R_{\text{max}}]$ and \textbf{goto} Step 1.
\item If $p_{\text{lower}} > p_{\text{upper}}$ then we use the classification $\mathcal{C}_{\text{lower}}$ and $\mathcal{C}_{\text{upper}}$ otherwise. Update the interval $[R_{\text{min}}, R_{\text{max}}]$ based on this classification and \textbf{goto} Step 1.
\end{enumerate}
\end{tcolorbox}

If the binary classification was perfectly accurate, then this algorithm would factor $\mathcal{N}$ using $\mathcal{O}\left(\log_2\left(\mathcal{N}\right)\right)$ binary classification models. The probability of this algorithm successfully factoring $\mathcal{N}$ would be approximately $\text{Pr}\left(\text{success}\right) \approx {\overbar{p}}^{\log_2\left(\mathcal{N}\right)/2}$, where $\overbar{p}$ is the mean out--of--sample binary classification accuracy of the $\log_2\left(\mathcal{N}\right)/2$ trials. For example, consider the currently unfactored RSA--260 semiprime with a greatly improved (and currently unattainable) binary classification probability of $\overbar{p} = 0.975$, then the probability of a successful factorisation would be approximately $\text{Pr}\left(\text{success}\right) \approx 1.8\times 10^{-5}$, which is not great, but non--zero. \\

It may be possible to improve the algorithm by the use of backtracking -- if at some point we arrive at an interval where both $\mathcal{C}_{\text{lower}}$ and $\mathcal{C}_{\text{upper}}$ are \texttt{False}, then we backtrack and build models to classify the corresponding opposing intervals\cite{karp2007}. 

\section{Discussion \& Future Directions}

We have described an interesting result from a deep learning--based binary classification experiment of large semiprimes. Our intuition was that the classification would be no better than a coin toss, however we found an out--of--sample accuracy of $0.72$. Thus, the binary classifier has some predictive capability concerning the ratio of the underlying primes of the semiprime. \\

As this result is unexpected, we encourage others to replicate this result. It may be possible that this is the result of a known, but obscure property of semiprimes which is not known to the author. \\

Unless the binary classification model is substantially improved then it is unlikely that this algorithm will result in a scalable, practical factorisation algorithm. However, given the recent advances in machine learning, perhaps it is possible for this algorithm to be improved to a point where it can quickly factor large semiprimes. \\

There are many future experiments we would like to perform, including: 

\begin{itemize}
\item Training multiple feature models. Presently we have used the binary representation of $\mathcal{N}$, however (with the trade-off of significantly more memory) we could train a multi--feature model with many different base representations. 
\item Train a large number of individual models with different base representations of the training data, then create an ensemble classification model. 
\item Does the out--of--sample accuracy improve if a larger dense ANN is used? Similarly for LSTM and  Transformer--based models? 
\item Does the out of sample accuracy improve if we greatly increase the number of semiprimes used in the training data? At present we do not have the memory required to train larger models. 
\item Can we reverse engineer the model and construct an algorithmic approach to the classification problem? 
\end{itemize}

\section{Code}

All the code in this paper and many additional experiments can be found on github\cite{deep_factor}. 


\bibliographystyle{abbrv}

\end{document}